\documentclass[sigconf]{acmart}
\usepackage{tabularx}
\usepackage{comment}

\copyrightyear{2026}
\acmYear{2026}
\setcopyright{cc}
\setcctype{by-nc-nd}
\acmConference[AI Summit '26]{ACM AI Summit 2026}{August 31-September 02, 2026}{Atlanta, GA, USA}
\acmBooktitle{ACM AI Summit 2026 (AI Summit '26), August 31-September 02, 2026, Atlanta, GA, USA}
\acmDOI{10.1145/3806096.3844885}
\acmISBN{979-8-4007-2638-5/2026/08}

\begin{document}

\title{From Protocols to Evidence: Bounded Claims for AI in Service of the Common Good}

\author{Nitesh V. Chawla}
\affiliation{%
  \institution{University of Notre Dame}
  \city{Notre Dame}
 % \state{Indiana}
  \country{USA}
}
\email{nchawla@nd.edu}

\author{Paolo Benanti}
\affiliation{%
  \institution{{LUISS} Guido Carli University}
  \city{Rome}
  \country{Italy}
}
\email{pbenanti@luiss.it}

\renewcommand{\shortauthors}{Chawla and Benanti}

\begin{abstract}

Claims that artificial intelligence (AI) systems improve decisions, broaden access, reduce harm, or empower users can exceed what their evaluation supports. This gap motivates our ongoing work on RISE AI. In this paper, we examine the proposed framework in light of Pope Leo XIV's \emph{Magnifica Humanitas}, particularly its treatment of human dignity, ownership, and technological power.

We develop three linked propositions. First, AI adoption can reveal unmet human needs and institutional weaknesses, while deployment can change how institutions respond to them. We propose a rupture test to examine whether deployment repairs or compounds existing failures, substitutes for human support, or conceals problems behind improved performance. The test would compare deployment with existing practice and feasible human or non-AI alternatives. Second, responsible AI must examine the institutional conditions surrounding a system, including who owns the infrastructure, controls its use, and bears responsibility for its consequences. Third, favorable evaluation cannot by itself establish respect for human dignity or moral and political legitimacy. We distinguish evidence-bounded deployment, which limits claims to what the evidence supports under specified conditions, from measurement-bounded governance, which identifies constraints that favorable results cannot override. 
We outline a proposed RISE AI framework for applying these propositions across Responsibility, Inclusivity, Safety, and Empowerment.
 %Institutions would state each claim, identify its supporting evidence and the authority responsible for acting on it, and specify when it should be narrowed or withdrawn. The framework would also record the basis of deployment constraints and who has authority to interpret them.

\end{abstract}




\maketitle

\section{Introduction}

As institutions and individuals delegate tasks to AI, they must also judge the efficacy of those AI outputs 
~\cite{bucinca2021}, and retain the ability and authority to intervene~\cite{siebert2023}. 
 A system may perform its assigned task well while weakening access to human expertise, making decisions harder to challenge, or obscuring who answers for harm~\cite{selbst2019,raji2020}. Evidence collected for one population or system version may not support the same claim in another. Evaluation must examine these effects and limits alongside technical performance. It must also consider the conditions that preceded adoption of AI, as consequences of AI also take shape through the people and institutions that develop, deploy, and use it.
 
 These conditions matter when people turn to AI for guidance and support. AI's availability and responsiveness raise questions about what people are seeking and what their institutions or relationships may have failed to provide. % to meet, and what institutions and relationships have failed to provide? 
%AI can be available, responsive, and patient in ways that institutions and relationships are not. Its adoption raises questions about what people need and what their institutions have failed to provide.
If a technology that reached its current scale only recently can so quickly unsettle our understanding of what it means to be human, then the problem may not be the technology alone. %Some of the needs that AI is being asked to meet existed well before its adoption, and AI can be a diagnostic for those failures. %, and AI has surfaced those. How people use it may help us recognize those needs, although adoption alone does not establish institutional failure. % We have to examine the failures that have always existed in society, and lean on AI as a diagnostic about those failures that AI is asked to fill in for as it can unravel needs that existing institutions did not fully meet. An institution's existing failures also do not excuse harms introduced by AI, but they matter when judging whether a deployment counts as improvement.
%It asks us to examine where institutions had already begun to reduce intelligence to performance, education to what can be measured, relationships to transactions, and human worth to productivity.
In this sense, AI can be both revelation and intervention. Its use may expose failures of care, belonging, and accountability; its deployment can change how institutions respond to them. An institution may use AI to address a failure, deepen it, replace human support, or conceal an unresolved problem behind better performance measures.  To judge whether AI improves a service, we need to understand where that service was already failing and whether deployment addresses those failures. Any new harms introduced by AI still require scrutiny, even when the previous service was inadequate.

%Over the last decade, 

AI ethics has articulated commitments to fairness, accountability, safety, and human oversight~\cite{jobin2019,sadek2025,bach2025, ruster2025}. These commitments increasingly appear in standards, legal requirements, and assurance practices~\cite{nist2023rmf,iso2023,eu2024ai,sabuncuoglu2025}. Their interpretation depends on context, and contemporary technologies and institutions may still lack adequate mechanisms for translating them into system requirements, organizational responsibilities, and evaluation practices. Where such mechanisms exist, institutions must establish whether they fulfill those commitments and what evidence supports that judgment.

%AI ethics has accumulated an expanding vocabulary of fairness, transparency, accountability, safety, human oversight, non-maleficence, and related commitments~\cite{jobin2019,sadek2025,bach2025}. However, this growing number of principles has not brought equal clarity what those principles require in practice or in context or what evidence would show that they have been met~\cite{sadek2025,bach2025, ruster2025}. Across philosophical, civic, religious, legal, and human-rights traditions, societies have long stressed commitments of dignity, responsibility, compassion, truthfulness, care for the vulnerable and common good. Many of these commitments also now appear in standards, organizational missions, assurance practices, and  legal requirements~\cite{nist2023rmf,iso2023,eu2024ai,sabuncuoglu2025}. The AI challenge is not that humanity lacks values, rather it is that contemporary technologies and institutions often lack mechanisms for translating those values into system requirements, organizational roles, accountability mechanisms, evaluation protocols, and deployment practices. %The question for this paper is therefore not simply how principles become practice, but what the resulting protocols and their evidence actually establish.

Pope Leo XIV’s Magnifica Humanitas informs our treatment of dignity, technological power, and the common good. It directs attention beyond the artifact to the institutional, economic, and political order in which AI is developed and used~\cite{leo2026magnifica}. Drawing on this frame, we examine what constitutes improvement and which uses remain unacceptable regardless of measured performance. %The challenge is to develop the algorithmic, engineering, and institutional innovations through which human commitments become operational in AI systems, just as they have through technological innovations over time. 

We develop three linked propositions. 
First, AI adoption can reveal unmet needs, while deployment can change how institutions respond to them. We propose a rupture test to assess those changes against existing practice and feasible alternatives.
Second, responsible AI must examine the ownership and decision authority surrounding a system, including who bears the consequences of its use. Third, evidence of improved performance cannot by itself justify a deployment or establish respect for human dignity.

We propose RISE AI as a framework for applying these propositions to claims about responsibility, inclusivity, safety, and empowerment. Institutions would identify the evidence supporting each claim, who is responsible for acting on it, and the constraints that favorable results cannot override. %This paper sets out the framework’s conceptual basis and proposed structure; implementation and empirical validation remain future work.

\section{AI, Human Dignity, and the Social Order}
%\vspace{-0.1in}
%\paragraph{\emph{Magnifica Humanitas} and the Social Order.}

Pope Leo XIV's \emph{Magnifica Humanitas} frames AI as one of the \emph{res novae} of our time, in continuity with the social transformations addressed by \emph{Rerum Novarum} in 1891~\cite{leo2026magnifica,leo1891rerum}. %We treat the encyclical as a moral and anthropological frame, not as a technical specification.
%It places those economic, institutional, and political structures within a moral account of human dignity, technological power, responsibility, and the common good.
%It asks us to look beyond the behavior of an AI system to the economic, institutional, and political structures through which technology is conceived, financed, controlled, and used.
%It enlarges the object of responsible AI from the behavior of an artifact to the economic, institutional, and political order in which technology is conceived, financed, controlled, and used. 
%RISE is placed in conversation with this horizon, not alongside it as a competing framework or as its technical implementation.
%RISE addresses only one part of this broader account. It is neither a competing moral framework nor a technical implementation of the encyclical.
%Chapter Five places technology within a broader \emph{culture of power}, the normalization of war, autonomous weapons, and the crisis of multilateralism (\emph{Magnifica Humanitas}, paras.~188--224). Earlier, the universal destination of goods is extended to patents, algorithms, platforms, technological infrastructure, and data (para.~67). 
The encyclical extends the universal destination of goods to patents, algorithms, platforms, infrastructure, and data (para. 67), observes that control over platforms, infrastructure, data, and compute often lies with major economic actors that set conditions of access and participation (para.~95), and warns that AI can amplify the power of those who already possess resources, expertise, data, and regulatory influence (paras.~106--110)~\cite{leo2026magnifica}. 
Technological innovations, including AI, are not neutral as they can foster participation and justice or intensify inequality, control, and exclusion (para.~85). Systems reflect the assumptions of those who design and train them (paras.~104, 111), while responsibility extends from developers and deployers to institutions, regulators, and users (paras.~105--111, 170, 209). Dignity is not one value to be optimized alongside others. It is an inalienable premise governing the legitimacy of institutions and the treatment of every person.

The encyclical also moves from these broader concerns to concrete action. Paragraph~14 connects dignity and the common good to responsible planning, human and social impact assessment, inclusion of vulnerable populations, and digital literacy. Paragraph~156 calls for verifiable measures protecting employment, retraining, and worker participation when AI is introduced. Paragraph~164 calls for consequential algorithmic decisions to be understandable, contestable, and subject to oversight~\cite{leo2026magnifica}. These passages raise practical questions for system design: can people understand and challenge a decision, and can responsibility for it be traced? Protocols can help establish whether these protections work in practice. Judgments about whether a particular use of AI is acceptable still require moral and political reasoning.

\section{From Commitments to Evidence}

%\paragraph{From Commitments to Protocols.}

Commitments to human agency and accountability have practical implications for how institutions design and use AI. Laws, standards, and assurance practices already give these commitments form through requirements for oversight, documentation, and review. The following chain traces how a commitment can inform system design and evaluation, ending with a claim whose scope is supported by the evidence.

\begin{quote}\small\raggedright
\textbf{Commitment} $\rightarrow$ \textbf{Design Objective} $\rightarrow$ \textbf{System Requirement} $\rightarrow$ \textbf{Implementation Mechanism} $\rightarrow$ \textbf{Evaluation Protocol} $\rightarrow$ \textbf{Bounded Evidence Claim}.
\end{quote}

%The chain begins with the rupture test discussed in Section 1: what institutional or relational failure already exists, what human or non-AI alternative is available, and which human capability is at risk of substitution? The resulting design should then be tested for whether it repairs, compounds, substitutes for, or conceals that failure. This baseline matters especially for empowerment. A claim that a system empowers users is incomplete unless we ask: compared with what alternative, for which users, according to whose definition of agency, and at what cost to others?

%Claims of improvement require comparison with existing practice and feasible alternatives. An inadequate status quo should not be the only benchmark.

The chain begins with a commitment the institution seeks to uphold. A design objective states what that commitment requires in the setting, and system requirements specify what must be built or put in place. Implementation mechanisms give those requirements practical form. An evaluation protocol then specifies how their operation and effects will be assessed. The findings determine which claims the institution can support and under what conditions.

Consider an institution using AI to recommend whether someone qualifies for a service. If the commitment is to preserve human agency, a design objective is to enable affected people to understand and challenge adverse recommendations. This requires an explanation they can use and access to a reviewer with authority to change the decision, without penalty for seeking review. The institution would need to record the basis of the recommendation, route challenges to that reviewer, and prevent irreversible action while review is pending. Evaluation would examine whether people understand the explanation, can complete the appeal, and obtain correction when warranted. It would also investigate who abandons the process or cannot access it. If the findings support a claim of meaningful control, that claim must remain limited to the decisions, populations, and conditions actually studied.

%Consider human agency in consequential AI-mediated decisions. A design objective may be that affected people retain meaningful control. System requirements may include disclosure of AI use, understandable reasons, alternatives, override and appeal channels, human review with authority, and protection against penalty for contesting a decision. Implementation mechanisms may include review interfaces, versioned decision provenance, escalation workflows, and constraints against irreversible automated action. Evaluation then requires more than a usability test: it may combine comprehension studies, appeal and override logs, review outcomes, subgroup analyses, and audits of whether alternatives are practically available. Only then can the institution justify a bounded claim: for a specified system version, class of users, decisions, channels, and time period, the system provided meaningful control under stated conditions.

This translation changes the unit of responsible AI from the model to the sociotechnical system (institution) itself. 
An accurate model can participate in an irresponsible system if no actor answers for failure. A model that appears safe under benchmark evaluation can become unsafe through workflow integration, user overreliance, absent recourse, or weak institutional capacity. A usable interface can still disempower users who cannot understand, contest, override, or refuse its recommendations.
\vspace{-0.1in}
\paragraph{Design Patterns.} Existing pattern catalogues describe ways to implement governance objectives in AI systems~\cite{lu2024patterns}. We draw on this work to discuss mechanisms for meeting those objectives and the evidence needed to assess whether they work in practice.\\
%Design patterns connect governance objectives to engineering practice. We do not claim that the following patterns are individually new; related governance, process, and product patterns have been catalogued elsewhere~\cite{lu2024patterns}. Their purpose here is to connect a governance objective to an implementable mechanism and then specify the evidence needed to determine whether it works in context. 
\noindent 1) Answerability-by-Design. Consequential systems should specify who answers for which decision at each lifecycle stage through role definitions, provenance, audit logs, escalation, and redress. The rupture test asks whether AI restores institutional answerability or further diffuses it across vendors, deployers, and users.\\
\noindent 2) Contestability-by-Design. Affected people need understandable grounds for challenge, usable appeal or refusal pathways, timely human review with authority, and protection against retaliation. Evaluation must test practical use, not merely the presence of an appeal button.\\
\noindent 3) Agency-Preserving Interfaces. Interfaces should expand the capacity to understand options and act on self-determined goals. Useful friction and cognitive forcing functions can reduce overreliance on AI, while meaningful human control requires that people retain the ability and authority to act on their responsibilities~\cite{bucinca2021,siebert2023}.\\
\noindent 4) Context-Aware Evaluation. Benchmark results do not constitute universal evidence. Evaluation must match the deployment population, workflow, institution, and version. 
A general score may support a narrow capability claim without supporting a claim about its effects in a particular setting.\\
\noindent 5) Evidence-Bounded Deployment. 
Deployment claims should not exceed what the evidence supports.
Evidence gaps, conflicts, and expiry conditions should be recorded explicitly. They should trigger qualification of the claim, additional evaluation, remediation, or withdrawal when necessary.

%Across these patterns, the same question recurs: what changes when AI enters the institutional or relational setting? Where a rupture already exists, does the system repair it, compound it, substitute for the human capability that should address it, or conceal it behind improved performance?  A system can satisfy these technical requirements and still violate a categorical constraint or operate within an institutional order whose legitimacy those requirements do not establish.% We develop that boundary below.

\vspace{-0.1in}
\paragraph{EU AI Act Requirements and Mechanisms.}

The EU AI Act translates governance commitments into requirements for AI development and use. For high-risk systems, Articles 9–15 establish technical and organizational requirements, including documentation, logging, and human oversight. Articles 27 and 43 address fundamental rights impact assessments for specified deployers and conformity assessment, while Articles 72–73 cover post-market monitoring and serious incident reporting~\cite{eu2024ai}.

Evaluation must examine how these requirements work in practice. Article 26 requires oversight personnel to have the necessary competence, training, authority, and support; evaluation must establish whether they can exercise that authority in their working conditions~\cite{eu2024ai}. A log may show that an appeal was processed without showing whether the person obtained meaningful review or correction. Compliance records can support these inquiries, but their existence alone does not establish that a deployment empowers people or addresses an institutional failure.

%The EU AI Act is one of the most comprehensive legal attempts to translate high-level commitments into an operational governance chain. 
%For high-risk systems, Articles 9–15 establish requirements concerning risk management, data governance, documentation, logging, information for deployers, human oversight, accuracy, robustness, and cybersecurity. Article 27 requires specified deployers of certain high-risk systems to conduct fundamental rights impact assessments, while Article 43 establishes conformity-assessment procedures. Articles 72–73 extend governance into post-market monitoring and serious incident reporting~\cite{eu2024ai}. 

%However, an oversight requirement does not tell us whether the reviewer has the time, authority, understanding, and institutional protection needed for meaningful control. A log does not establish which claim it supports or whether an appeal produces correction. The Act gives responsible AI a legal and institutional framework. But compliance artifacts do not by themselves establish empowerment or institutional repair, and they cannot establish the legitimacy of the surrounding distribution of technological power. 
Our proposal for RISE AI focuses on what claims those records support, under what conditions, and where that support ends. 
%RISE AI addresses the narrower evidentiary question: what claims those artifacts support, under what conditions, and where that support ends.
%Compliance artifacts can support broader claims, but they cannot establish empowerment, institutional repair, or a just distribution of technological power.The Act gives responsible AI a legal and institutional framework. But compliance alone does not establish empowerment, institutional repair, or a just distribution of technological power. That is the narrower problem an evidence architecture such as RISE is intended to address.

\section{Limits of Design and Measurement}

% This section adapts the contribution supplied by Paolo Benanti. The
% highlighted review copy marks the full section in purple.

%Two limits remain even when governance is well designed, and neither can be resolved by better measurement alone.

%\vspace{-0.1in}
%\paragraph{Political Economy and System Boundaries.}

%The shift from the model to the sociotechnical system is a real advance, but the sociotechnical system is not the largest relevant unit. 

Evaluating the sociotechnical system brings institutional practices into view, but the scope of evaluation still matters. %Evidence that users can challenge decisions may leave other questions unanswered: who owns the infrastructure, who determines the system’s purpose, and whose labor and resources sustain it~\cite{crawford2021atlas}.
%The shift from the model to the sociotechnical system is an important advance as it can begin to capture the deployment context. However, evaluation may still not offer a lens into the governance of such a system.  
An evaluation methodology may establish that a deployed system preserves meaningful control for its users, offers practicable contestation, and communicates uncertainty well. At the same time, it may say nothing about who owns the compute on which it runs, who set the agenda under which it was built, or whose labor and resources sustain it~\cite{crawford2021atlas}.

\emph{Magnifica Humanitas} makes these questions part of the object of governance. It treats algorithms, platforms, infrastructure, and data as goods whose concentration can violate their universal destination; identifies private control over data and compute as a source of political and economic asymmetry; and asks who can train and govern systems and who is merely subjected to them (paras.~67, 95, 106--110)~\cite{leo2026magnifica}. %A validity chain that terminates at the deployment context can therefore certify a well-designed system inside an unjust distribution.here
Evidence supporting a claim about a deployment does not by itself establish that the surrounding distribution of power is just. Ownership, labor, resource use, and regulatory influence therefore remain subjects of governance.
%A validity chain can support a bounded claim about a system within its deployment context while saying nothing about whether the larger distribution of power is just, and thus measurement becomes limiting as part of a broader governance claim.
%Measurement of the artifact is not measurement of the order the artifact serves. 
%Governance must therefore continue to consider attributes such as ownership,  labor, resources and regulatory influence even when they cannot be reduced to system-performance indicators.

%\vspace{-0.1in}
%\paragraph{Dignity and Construct Validity.}

{\it{Construct validity}} concerns whether observed indicators and empirical relationships support an intended interpretation of a construct~\cite{cronbach1955,messick1995,jacobs2021}. We treat inherent human dignity as a moral premise. Observable conditions may provide evidence that dignity has been respected or violated, but they do not measure how much dignity a person possesses or determine whether its violation is justified by other benefits. A system can perform well on every selected measure and still treat a person merely as a means.

Conditions relevant to agency, recourse, and access can change when AI is introduced. People may gain access to a service while losing the ability to challenge its decisions or obtain human assistance. Within RISE AI, such claims would be assessed against evidence from the population and setting studied. The validity question is whether the evidence supports the claim being made; a favorable result does not establish that the deployment respects human dignity. %Human worth and dignity do not vary with system performance; no indicators can establish, increase, or offset them. By contrast, conditions relevant to agency, recourse, or access may vary across populations, institutions, workflows, and system versions. %RISE AI can evaluate whether specified evidence supports bounded claims about variable conditions.

Evidence-bounded deployment requires institutions to limit their claims about an AI system to what the evidence supports under the conditions evaluated. We use \textit{measurement-bounded governance} to describe a further limit: favorable results cannot justify every use of a system. Some uses remain unacceptable even when performance, such as accuracy, is strong. An institution may, for example, require access to human review regardless of gains in processing speed. An evidence profile should therefore record these constraints alongside the evaluation results, making clear which commitments those results cannot override.

These commitments set constraints on deployment. They are not constructs to be measured, although evidence may show whether they have been breached. For each constraint, the governance record should state its basis, whom it protects, and who has authority to interpret or revise it. That basis may be law, institutional policy, or a commitment made to an affected community. Article 5 of the EU AI Act provides an example: where a practice is prohibited, favorable performance does not make it permissible~\cite{eu2024ai}. Table~\ref{tab:measurement-boundary} summarizes the distinction between claims assessed through evidence and constraints that the evaluation must respect.

\begin{table}[t]
\caption{Evidence-bounded claims and measurement-bounded constraints.}
\label{tab:measurement-boundary}
\scriptsize
\begin{tabularx}{\columnwidth}{@{}p{0.15\columnwidth}XX@{}}
\toprule
 & \textbf{Evidence-bounded claims} & \textbf{Measurement-bounded constraints} \\
\midrule
Question & What does evidence support here? & What may not be authorized? \\
Object & Variable conditions: agency, recourse, access, safety, answerability. & Dignity-based constraints, legitimate refusal, prohibited uses, non-substitutable relationships. \\
Evidence & Supports, qualifies, or contradicts a claim. & May reveal a breach; cannot override the boundary. \\
Basis / authority & Predeclared protocol and accountable decision process. & Law, policy, affected communities, institutional mission, moral judgment. \\
Decision & Support, qualify, remediate, retest, or withdraw. & Prohibit, stop, redesign, or require a human alternative. \\
Proposed RISE role & Evaluates and records evidentiary support. & Records the boundary, basis, authority, and revision process. \\
\bottomrule
\end{tabularx}
\end{table}

Responsible AI requires both better design and continued moral and political judgment. Moral and political judgment remains necessary because AI-mediated systems can redistribute agency and responsibility, raising questions of authorship, accountability, and authority that design alone cannot settle. %AI does not demand better design in place of moral and political judgment. %It demands both. %The engineering and moral-political agendas must move forward together.
%Technical benchmarks cannot establish moral legitimacy, just as ethical ideals cannot substitute for system design.

%This sharpens the paper’s central claim. AI does not demand better design in place of moral and political judgment. It demands both. %Better design is required because moral commitments need reliable purchase on systems that act; continued moral development is required because delegated and distributed agency raises questions about authorship, responsibility, and power that inherited categories do not fully resolve.
%Better design is necessary because moral commitments must shape how systems actually behave. Moral and political judgment remains necessary because AI redistributes agency and responsibility, raising questions of authorship, accountability, and power that design alone cannot settle. The engineering and moral-political agendas must move forward together. 
%Technical benchmarks cannot establish moral legitimacy, just as ethical ideals cannot substitute for system design.
%Neither waits for the other, and neither can authenticate the other. Any evidence architecture must remain bounded by both.

\section{RISE AI as an Evidence Architecture}

%The encyclical provides a moral and anthropological frame for understanding AI in relation to dignity, political economy, and the common good. RISE AI addresses a narrower operational question: when institutions make claims about a particular AI system, what evidence supports those claims, and where should those claims stop?

 The proposed RISE AI framework would organize bounded system claims around  the four constructs of Responsibility, Inclusivity, Safety, and Empowerment, answering four questions: Who answers for the system? Whose perspectives shape it and whose needs does it serve? Whom does it protect from harm? Who gains or loses agency through it? 
We propose the following validity chain for assessing each claim.

\begin{quote}
\textbf{Construct} $\rightarrow$ \textbf{Bounded System Claim} $\rightarrow$ \textbf{Indicator} $\rightarrow$ \textbf{Evidence Source} $\rightarrow$ \textbf{Bounded Inference}.
\end{quote}

A claim that AI improves a service or gives people greater control requires a clear comparison with existing practice and feasible alternatives. Within RISE AI, the question would be whether the evidence supports that claim under the conditions studied. A metric is useful only if there is a defensible connection between what it measures and what the institution claims~\cite{cronbach1955,messick1995,jacobs2021}.

Our proposal builds on existing governance and assurance work. NIST AI RMF, ISO/IEC 42001, and the EU AI Act provide guidance, requirements, and obligations that institutions must interpret for their settings~\cite{nist2023rmf,iso2023,eu2024ai}. Pattern catalogues describe practices for implementing governance objectives~\cite{lu2024patterns}, while assurance cases connect claims to arguments and evidence~\cite{sabuncuoglu2025}. Documentation, provenance records, and operational logs can contribute to that evidence~\cite{werder2022,foalem2025,sabuncuoglu2025}. RISE AI would use these resources to examine what an institution can reasonably claim about a deployment and when changes to the system or its use require that claim to be reviewed.

%Where the claim is comparative, such as a claim of improvement or empowerment, the chain must also specify the relevant baseline and feasible alternatives. Accordingly, RISE evaluates whether the evidence supports the bounded inference rather than treating a metric as inherently valid~\cite{cronbach1955,messick1995,jacobs2021}. RISE does not replace existing governance frameworks. NIST AI RMF provides risk-management guidance, ISO/IEC 42001 specifies management-system requirements, and the EU AI Act establishes legal obligations~\cite{nist2023rmf,iso2023,eu2024ai}. Responsible AI pattern catalogues supply reusable governance, process, and product practices~\cite{lu2024patterns}. Assurance cases structure claims, arguments, and evidence~\cite{sabuncuoglu2025}. Model and data documentation, provenance records, and operational logs provide candidate evidence~\cite{werder2022,foalem2025,sabuncuoglu2025}. A standard supplies requirements; a pattern supplies a candidate mechanism; an operational system supplies an artifact; RISE AI records what claim the artifact supports and where that support stops.

%The primary artifact is a versioned claim-evidence graph, not a universal governance score. 
We propose a versioned claim-evidence graph to connect claims with their supporting evidence. An evidence profile would present the relevant part of that graph for a particular system, context, and evaluation period. The graph would link each bounded claim to evidence artifacts, scope conditions, accountable actors, and constraints,  
recording evaluators’ judgments about whether the evidence supports, qualifies, or contradicts the claim.
At minimum, the record should identify the system, model, data, and interface version; the relevant population and context; the accountable owner; the indicator and threshold; the provenance and limitations of the evidence; and any expiry or change trigger. For each claim, the proposed record would distinguish whether the evidence is direct or proxy-based, whether it supports or conflicts with the claim, and whether required evidence is missing or stale. A \emph{context-and-power record} would document ownership, financing, and decision authority, along with the labor and resources the system depends on and who bears its risks. This would allow reviewers to examine whether those arrangements limit the institution’s ability to change or withdraw the system.

%We also propose linking each claim-evidence graph to a \emph{context-and-power record}. This record would identify who owns and controls the system and its infrastructure, who finances it, and who decides its purpose and terms of use. It would also document the labor and resources needed to sustain it and how risks and benefits are distributed. 
%Reviewers could then consider, for example, whether a vendor dependency limits an institution's ability to change or withdraw the system. The record would support scrutiny of these arrangements without assigning them a single score or treating favorable system performance as evidence that they are just.

%Each graph is also linked to a \emph{context-and-power record}: ownership and control of models, data, compute, and platforms; financing and procurement relationships; labor and resource dependencies; agenda-setting authority; and the distribution of risks and benefits. These are not additional indicators from which a political-economy score is calculated.Their purpose is to keep the institutional setting visible when a system-level claim is evaluated. RISE AI cannot determine through measurement whether that institutional order is just. It can, however, make clear that system-level evidence does not establish the legitimacy of that order.
\vspace{-0.1in}
\paragraph{Operational Protocol and Institutional Baseline.}

We propose an assessment process that institutions could use alongside their existing evaluation and review procedures. The team would first define the system, the people affected, and the problem the deployment is intended to address. It would document existing institutional conditions and identify feasible alternatives, including human and non-AI approaches. %The context-and-power record would capture ownership and decision authority.

The team would then formulate claims with affected communities, domain experts, and system owners, documenting disagreements about what should count as improvement. Before evaluation, it should specify the evidence needed to assess each claim, the criteria for judging the findings, and the constraints on deployment. Evidence could come from system tests, user studies, or records of decisions and appeals, depending on the claim. The assessment would distinguish direct observations from proxy measures and report conflicting findings, missing evidence, and results that are no longer current. Changes to the system or its institutional setting would prompt review of the affected claims and the deployment or procurement decisions based on them.

What counts as timely review or sufficient understanding depends on the setting. Under the proposed framework, institutions would explain their thresholds, identify who approved them, and provide a way to challenge those choices. A claim could take the following form: \emph{For this system version, population, setting, and period, the evidence supports [claim] under [conditions]. It leaves [questions] unresolved. Changes to [specified conditions] would require the claim to be reviewed.}

The proposed rupture test would examine what improves and what may be lost when AI is introduced. Faster processing, for example, could reduce delays while cuts in caseworker availability make it harder for applicants to obtain help. Comparison with existing practice and feasible alternatives would need to account for both effects. An institution claiming to have addressed a service failure would therefore need evidence beyond processing speed, including whether people can still obtain assistance and challenge a decision.

\vspace{-0.1in}
\paragraph{Public-Benefits Scenario.}
Suppose an AI system supports public-benefits eligibility decisions in an institution already marked by slow processing, limited caseworker capacity, and difficult appeals. Faster decisions would not, by themselves, show that applicants have better access to assistance or appeals. Table~\ref{tab:contestability} shows how one RISE chain would apply to an empowerment claim based on meaningful contestability of an adverse recommendation.

\begin{table}[htp]
\caption{A compact RISE evidence contract for contestability.}
\label{tab:contestability}
\small
\begin{tabularx}{\columnwidth}{@{}p{0.24\columnwidth}X@{}}
\toprule
\textbf{Element} & \textbf{Public-benefits instantiation} \\
\midrule
Claim & Applicants can meaningfully contest adverse AI-mediated recommendations. \\
Requirements & AI disclosure; understandable reasons; accessible online and offline appeal; authorized human review; non-retaliation; correction of downstream effects. \\
Mechanisms & Decision notice; reason codes; appeal endpoint; review queue; adverse-action pause; correction and recovery workflow. \\
Evidence & Versioned decision, notice, appeal, review, incident, complaint, and recovery logs; accessibility and comprehension studies. \\
Indicators & Notice comprehension; appeal initiation and completion; abandonment; review latency; reversal; subgroup disparities; recovery completeness. \\
Boundary & Specified system version, jurisdiction, channels, languages, populations, decisions, and evaluation period. \\
Context~\&~power & System and data owner; vendor and cloud dependencies; procurement terms; eligibility-policy authority; caseworker staffing effects; control of appeal records. \\
Constraints & No irreversible adverse action without authorized human review; no online-only appeal; no retaliation for contesting a decision; preserve practicable access to a caseworker. \\
\bottomrule
\end{tabularx}
\end{table}

%Logs can trace decisions and corrections, but a claim of meaningful contestability would also require evidence that applicants can use the appeal process and reviewers have authority to act. The claim should identify groups or channels not covered by the evaluation, such as phone-only applicants.

%Logs support audit and recovery but do not establish contestability without user studies, case-file audits, subgroup coverage, and human reviewers with practical authority. The resulting claim must state what is supported and what remains unknown, such as phone-only applicants, additional languages, or another jurisdiction. The context-and-power row keeps ownership and institutional capacity visible; the constraints row records conditions that favorable aggregate performance cannot offset.

\vspace{-0.1in}
\paragraph{Example Use Cases}

Education and clinical decision support offer settings in which the proposed framework could be studied. In education, a claim that AI empowers students would require evidence beyond task completion or correct answers. Can students apply what they have learned without the tool? Do they retain control over their work and access to a teacher when needed? In clinical decision support, evaluation would need to examine how clinicians use a recommendation, whether they can override it, and what happens when it is wrong. Predictive accuracy alone would not establish safety in practice. In either setting, any resulting claim would need to specify whom the evidence covers and the conditions under which it was collected. These examples identify questions for future studies; they do not establish that RISE AI is effective in either domain.

\section{Research Roadmap}

The next phase requires detailed operationalization, implementation, and empirical testing of RISE AI. Seven questions matter most. \emph{Participatory content validity} asks whether affected communities, domain experts, and accountable institutions judge bounded claims to cover what matters and where their definitions diverge. \emph{Inter-evaluator reliability} tests whether independent evaluators classify the same evidence, gaps, and inference boundaries similarly. \emph{Diagnostic value} examines whether RISE AI profiles reveal unsupported claims, evidence gaps, or failures missed by benchmarks and compliance review. \emph{Change sensitivity} asks whether claims are appropriately revised or invalidated when underlying data, models, deployment or use contexts change. %models, data, interfaces, workflows, populations, ownership, procurement, or institutional capacity change. 
\emph{Decision utility} asks whether evidence profiles improve the justification of procurement, deployment, and remediation decisions. \emph{Political-economic sensitivity} examines whether the context-and-power record exposes dependencies or distributions of risk and benefit that system-level assessment misses and whether those disclosures affect decisions. \emph{Operational burden and proportionality} examines the expertise, time, infrastructure, and cost required, including whether credible implementation remains feasible beyond large organizations.

Initial studies could use draft evidence profiles and manual review; testing the full process would require an implementation of RISE AI, which is part of future work. %The implementation will need to represent the rupture test, validity chains, evidence-bounded deployment, and measurement-bounded governance, while carrying explicit baselines and feasible alternatives, claim-evidence profiles, context-and-power records, constraints, provenance, and change triggers. 
This implementation should also sit alongside existing requirements repositories, audit systems, observability tools, and governance workflows rather than replace them. Building it will test whether the pieces of the framework actually work together, what burden they impose, and where the framework itself needs to change.

\paragraph{Limitations.}
The examples illustrate the proposal; they do not validate RISE AI. Claims, baselines, and thresholds depend on who participates and who has decision authority. An inadequate baseline can make a weak deployment appear beneficial. Evidence may underrepresent those most affected, and its collection can create privacy risks. Recording ownership and authority also does not ensure that either will be meaningfully scrutinized.

\section{Conclusion}

%AI governance is no longer confined to statements of principle. The EU AI Act, NIST AI RMF, ISO/IEC 42001, and assurance practices now provide legal obligations, management requirements, risk-management guidance, records, and procedures. None of this tells us, by itself, whether a particular deployment improves decisions, broadens access, reduces harm, or empowers people. Institutions still have to decide what the evidence supports, under which conditions, and what remains unresolved.

%Evaluation must address both the behavior of an AI system and the conditions in which it is used. Unreliable outputs and uneven performance can cause harm, as can overreliance, weakened recourse, and the displacement of human expertise. AI can also reveal failures of responsiveness, belonging, care, and accountability that preceded its adoption. Once deployed, it can repair those failures, compound them, substitute for weakened human capacities, or conceal them behind improved performance. The rupture test requires explicit institutional baselines and human and non-AI alternatives. It asks whether deployment strengthens the capabilities and relationships the institution is meant to support. Evaluation must also examine any new harms the deployment introduces.

Evaluation must examine how an AI system performs and what changes for people when an institution adopts it. Faster or more accurate decisions may still leave people with less access to human expertise or fewer ways to challenge an outcome. Some problems may predate adoption; others may result from the deployment itself. We propose a rupture test that would compare deployment with existing practice and feasible human or non-AI alternatives, helping examine whether the institution addresses the underlying problem and what harms or losses accompany any gains.

The proposed RISE AI framework brings these concerns into the assessment of institutional claims about AI. It identifies three boundaries: the comparison against which improvement is claimed, the scope of the supporting evidence, and the constraints that favorable results cannot override. The rupture test would address the comparison; validity chains and evidence-bounded deployment would define the scope of support; and measurement-bounded governance would make the constraints explicit.

Our reading of \emph{Magnifica Humanitas} connects these questions to dignity, technological power, and the common good. %Ownership, infrastructure, financing, labor, and political authority remain matters of governance even when a system performs well. Favorable evaluation results alone cannot establish the legitimacy of those arrangements. Nor can performance gains justify treating dignity as an outcome to be traded against other benefits. Dignity governs evaluation; it is not produced by it. 
The engineering and moral-political agendas must proceed together. Whether AI serves the common good depends on what people build and what institutions choose to protect. Those choices carry a responsibility to build systems and institutions worthy of humanity.

\end{document}